\pdfoutput=1

\documentclass[11pt]{article}

\usepackage[final]{acl}

\usepackage{times}
\usepackage{latexsym}
\usepackage{multicol}
\usepackage[T1]{fontenc}

\usepackage[utf8]{inputenc}

\usepackage{microtype}

\usepackage{inconsolata}

\usepackage{graphicx}
\usepackage{float}
\usepackage{subcaption}
\usepackage{caption}
\usepackage{appendix}
\usepackage{comment}
\usepackage{booktabs}
\usepackage{multirow}
\usepackage{adjustbox}
\usepackage{colortbl}
\usepackage{arydshln}
\usepackage{amsmath}
\usepackage{tabularx}
\usepackage{url}
\newcolumntype{Y}{>{\centering\arraybackslash}X}

\title{SiLQ: Simple Large Language Model Quantization-Aware Training}

\author{
 \textbf{Steven K. Esser\textsuperscript{1}},
 \textbf{Jeffrey L. McKinstry\textsuperscript{1}},
 \textbf{Deepika Bablani\textsuperscript{1}},
 \\
 \textbf{Rathinakumar Appuswamy\textsuperscript{1}},
 \textbf{Dharmendra S. Modha\textsuperscript{1}},
\\
\\
 \textsuperscript{1}IBM Research, San Jose, CA, USA
\\
 \small{
   \textbf{Correspondence:} \href{mailto:sesser@us.ibm.com}{sesser@us.ibm.com}
 }
}

\begin{document}
\maketitle
\begin{abstract}

Large language models can be quantized to reduce inference time latency, model size, and energy consumption, thereby delivering a better user experience at lower cost.  
A challenge exists to deliver quantized models with minimal loss of accuracy in reasonable time, and in particular to do so without requiring mechanisms incompatible with specialized inference accelerators. Here, we demonstrate a simple, end-to-end quantization-aware training approach that, with an increase in total model training budget of less than 0.1\%, outperforms the leading published quantization methods by large margins on several modern benchmarks, with both base and instruct model variants. The approach easily generalizes across different model architectures, can be applied to activations, cache, and weights, and requires the introduction of no additional operations to the model other than the quantization itself.

\end{abstract}

\section{Introduction}
Large language models (LLMs) have achieved remarkable capabilities across a wide range of AI tasks.  However, there are two major challenges emerging in wide-scale deployment of LLMs: energy consumption and response latency \cite{appuswamy2024breakthrough}. It is estimated that inference accounts for  80\% of the total energy cost of LLM solutions \cite{inferenceenergy2024}. In addition, current and emerging applications such as interactive dialog and agentic workflows require very low latencies. Both of these concerns are addressed by low precision neural inference processors, such as NorthPole \cite{modha2023neural}. Lower precision compute reduces power consumption directly, and by reducing area, allows model memory to be placed next to compute for further energy savings and lower latency \cite{modha2023neural}.

While nearly all LLMs are trained today using 16-bit precision, it is possible to quantize such models to run on low precision inference accelerators. To this end, a variety of techniques have emerged that seek to minimize quantization related accuracy loss, while providing fast time-to-solution, judged relative to total development time, simple implementation, and a quantized model that is fully compatible with the target deployment platform.  Efforts have focused on post-training quantization (PTQ), where quantization is tuned using a small amount of calibration data, or quantization-aware training (QAT), where a model is fine-tuned with differentiable quantization operators.  PTQ is argued to be preferable due to its low dataset and compute requirements, and recent work has shown it to outperform QAT based approaches \cite{liu2024spinquant}.

\begin{table*}
\begin{centering}
\begin{small}
\caption{SiLQ results in more accurate base and instruction-tuned LLMs than leading PTQ techniques across common sense reasoning and Open LLM benchmarks. The Bits column indicates bits to quantize activations, with "s" or "d" denoting static or dynamic quantization, then KV cache, then weights. SmoothQuant and SpinQuant results computed by us, using code published by each paper's authors.}  
\begin{tabular}{c|cc|cccccc}
\hline
\textbf{Model} &\textbf{Bits}  &\textbf{Method}  & \textbf{CSR} & \textbf{OLLM} & \textbf{OLLM} \\ 
\textbf{} & A-C-W  &\textbf{} & \textbf{} & \textbf{v1} & \textbf{v2}\\
\hline
\multirow{4}{*}{Llama-3-8B}
&\begin{tabular}[c]{@{}c@{}}16-16-16\end{tabular}
         & Baseline  & 67.09 & 62.65 & 13.70  \\
\cdashline{2-7}
& \begin{tabular}[c]{@{}c@{}}\multirow{3}{*}{8d-8-4}\end{tabular}
        & SmoothQuant*  &   58.73 & 45.15 & 6.65     \\
        && SpinQuant   &   63.97 & 57.99 & 12.28     \\
        && \cellcolor{gray!30}SiLQ   & \cellcolor{gray!30}67.20 & \cellcolor{gray!30}61.14 & \cellcolor{gray!30}12.66  \\ \hline
\multirow{4}{*}{\shortstack{Llama-3.1-\\Tulu-3.1-8B}}
&\begin{tabular}[c]{@{}c@{}}16-16-16\end{tabular} 
         & Baseline  & 69.91 & 69.56 & 26.45    \\
\cdashline{2-7}
&\begin{tabular}[c]{@{}c@{}}\multirow{3}{*}{8d-8-4}\end{tabular} 
        & SmoothQuant* & 64.20  & 58.12  & 20.49  \\
        && SpinQuant &   67.47 & 66.91 & 24.03   \\
        && \cellcolor{gray!30}SiLQ  & \cellcolor{gray!30}69.59  & \cellcolor{gray!30}69.79 & \cellcolor{gray!30}27.10 \\ \hline
\multirow{9}{*}{\shortstack{Granite-3.1-8B-\\Instruct}}
&\begin{tabular}[c]{@{}c@{}}16-16-16\end{tabular}
         & Baseline  & 68.46 & 72.11 & 29.91    \\
\cdashline{2-7}
&\begin{tabular}[c]{@{}c@{}}\multirow{3}{*}{8d-8-4}\end{tabular}
         & SmoothQuant* & 62.68 & 61.66 & 20.08     \\
        && SpinQuant &  65.96 & 65.52 & 21.35       \\
        &&\cellcolor{gray!30} SiLQ        &\cellcolor{gray!30}68.03 &   \cellcolor{gray!30}71.48 & \cellcolor{gray!30}29.14      \\
\cdashline{2-7}
&\begin{tabular}[c]{@{}c@{}}\multirow{2}{*}{8s-8-4}\end{tabular} 
        & SmoothQuant* & 49.06  & 35.24  & 8.93      \\
        &&\cellcolor{gray!30} SiLQ & \cellcolor{gray!30}67.41 & \cellcolor{gray!30}70.86 & \cellcolor{gray!30} 29.03       \\
\cdashline{2-7}

&\begin{tabular}[c]{@{}c@{}}\multirow{3}{*}{8d-4-4}\end{tabular}
         & SmoothQuant* &  56.48  &  39.78   &  7.15   \\
        && SpinQuant &    63.12 & 56.67 & 14.47     \\
        &&\cellcolor{gray!30} SiLQ  & \cellcolor{gray!30}67.92 & \cellcolor{gray!30}70.93 & \cellcolor{gray!30}29.13      \\
\hline
\multicolumn{3}{l}{*head not quantized} \\ 
\end{tabular}

\label{tab:ptq_all_evals}
\end{small}
\end{centering}
\end{table*}

Here, we provide a counterpoint to this perspective. We introduce a simple approach to QAT that requires increasing the total training budget by less 0.1\%, measured in training tokens, and that can make use of publicly available datasets or the model's original fine-tuning data.  The approach achieves accuracy several percentage points superior to the best published alternative quantization methods (Table \ref{tab:ptq_all_evals}) on zero-shot Common Sense Reasoning tasks (CSR)\footnote{As in \cite{liu2024spinquant}, see references therein} and the Huggingface Open LLM leader-board versions 1 and 2 (OLLMv1 and OLLMv2)\footnote{As in \cite{lee2024comprehensive}, see references therein}.

Our approach, Simple Large Language Model Quantization-Aware Training (SiLQ), is to: 1) add quantization to the model to match the target deployment configuration, using the straight through estimator \cite{bengio2013estimating} for training time gradients, 2) set quantizer step size values initially through calibration and then refine further using LSQ \cite{esser2019learned}, and then 3) train end-to-end with a standard workflow, allowing one to employ existing code frameworks, using knowledge distillation \cite{hinton2015distilling}, and either the model's original training dataset or a high quality public dataset.  We further introduce a weight calibration technique based on an approximation of mean squared error. We train on up to a few billion tokens, a small investment compared to the trillions of tokens used to train modern LLMs. 

\begin{figure}
\centering
\includegraphics[width=0.48\textwidth]{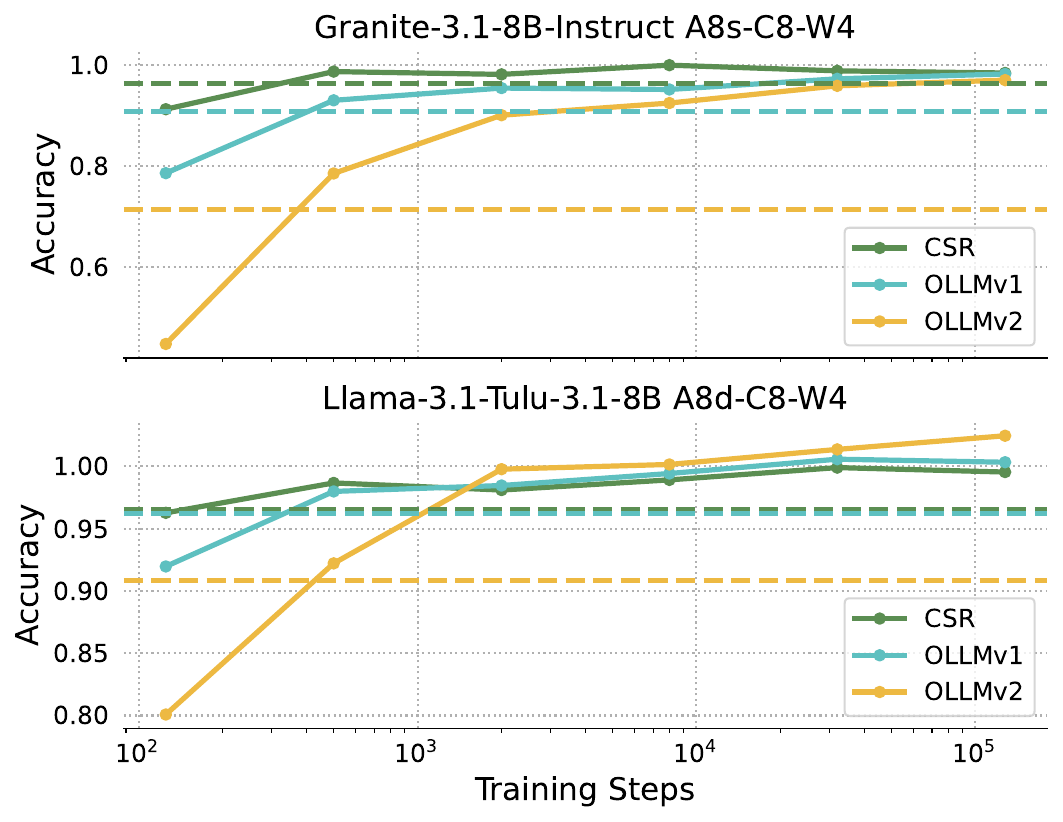}
\caption{Accuracy improves with longer QAT. The y-axis represents accuracy relative to the original fp16 model.  Horizontal dashed lines show PTQ method SpinQuant accuracy. Accuracy on the harder OLLMv1 and, in particular, OLLMv2 benchmarks improves the most with longer QAT, significantly outperforming PTQ.
}
    \label{fig:acc_vs_steps_rel}
\end{figure}

A benefit of SiLQ is solution scalability. To the limit of what is possible through deep learning, one can train longer to improve accuracy, which we demonstrate empirically (Figure \ref{fig:acc_vs_steps_rel}). This allows one to balance up-front training costs with deployment time performance as needed. In addition, QAT methods in principle offer better generalizability.  One simply adds precision constraints specific to a given accelerator, and lets the optimizer find the best solution tailored to those constraints. On the other hand, to reach acceptable accuracy, PTQ methods require identifying model-specific impediments to quantization, such as outliers appearing in particular layers, and then devising custom solutions for those issues \cite{frantar2022gptq, xiao2023smoothquant, liu2024spinquant, ashkboos2024quarot}.

Surprisingly few demonstrations have been published using QAT for typical LLM deployment scenarios. Existing work either does not quantize all tensors necessary for full acceleration \cite{liu2025paretoq}, targets extremely low precisions at the cost of accuracy degradation that would likely be unacceptable for most users \cite{du2024bitdistiller}, or introduces time-consuming complexities such as data self-generation \cite{liu2023llm}. Further, existing publications focus on quantizing the base version of various models, rather than the higher accuracy instruction-tuned version typically used in deployment.

We demonstrate our approach on models with an 8-bit activation, 4- or 8-bit cache, and 4-bit weight configuration. Remarkably, even when applied to instruction-tuned models that were originally created using a multistage process of pre-training, supervised fine-tuning, direct preference optimization,  proximal policy optimization, and model averaging \cite{lambert2024t}, our single stage end-to-end approach preserves accuracy at nearly the same level as the original model, outperforming all leading hardware-friendly LLM quantization methods.

\section{Related work}

A typical LLM consists of an initial embedding layer and final linear layer, the head, sandwiching a series of alternating attention and multi-layer perceptron blocks, each in turn comprising normalization, element-wise, matrix multiplication, and linear operations. Input is provided as a series of integer valued tokens, and the model stores intermediate responses to earlier tokens of an input sequence in  multiple caches.  The model is parameterized by various weights and biases learned during training. The linear and matrix multiplication layers comprise the majority of compute costs, and the cache and weights comprise the majority of memory costs. Both costs can be reduced by quantizing the cache and weights for storage, and all other inputs to linear and matrix multiplication layers, commonly referred to as activations, for compute.

In order to ground our work, we choose an inference accelerator, NorthPole \cite{modha2023neural}, and ensure that our quantization scheme is fully compatible with this deployment platform.  NorthPole employs a vector-matrix multiplication unit, used to compute linear and matrix multiplication layers, supporting any combination of 2-, 4-, \mbox{8-,} or 16-bit integer activations as operands, can store the cache and weights at 2-, 4-, or 8-bit, and uses fp16 precision for other operations.  We compare our approach to the three leading alternatives also compatible with these constraints, SmoothQuant \cite{xiao2023smoothquant}, SpinQuant \cite{liu2024spinquant}, and LLM-QAT \cite{liu2023llm}, but do not compare to approaches that would be unable to deploy to this platform.

SmoothQuant is a PTQ method that can effectively quantize LLMs with 8-bit weights, activations, and cache with no loss in accuracy. It increases the range of the weights in order to decrease the range of the activations, thereby reducing activation outliers, improving quantization accuracy.  Unfortunately, it performs poorly with 4-bit weights (Table \ref{tab:ptq_all_evals}) \cite{liu2024spinquant}.

SpinQuant is a state-of-the-art PTQ method that learns rotation matrices to reduce outliers, merges these rotations into the weights of the linear layers, then applies the GPTQ algorithm to quantize model weights. SpinQuant is shown to produce models with higher accuracy than methods with random weight rotations \cite{ashkboos2024quarot}. Surprisingly, it was also shown to be more accurate on a common-sense reasoning benchmark than the leading LLM QAT method discussed next.

LLM-QAT demonstrates superior results to the SmoothQuant PTQ method, and is compatible with our chosen constraints.   LLM-QAT uses approximately 100,000 self-generated training samples to finetune LLMs.  Data self generation removes dependency on external datasets, but requires a non trivial amount of compute time.

A number of other quantization methods exist that do not deliver a solution compatible with our chosen constraints, that is, that do not quantize all required tensors, or that do not reach acceptable accuracy in reasonable time, and so are not discussed beyond noting here.  Weight-only PTQ methods  like GPTQ \cite{frantar2022gptq} are commonly employed to reduce model size, and at 4-bit precision come within 1 to 2\% of the accuracy of full precision models \cite{lee2024comprehensive}.  EfficientQAT focuses only on making QAT efficient for large models, and uses weight-only quantization at 2- and 3-bit precision \cite{chen2024efficientqat}.  PrefixQuant shows a greater than two percent improvement in accuracy by adding QAT after PTQ, however it adds unquantized values in the cache which is not hardware friendly \cite{chen2024prefixquant}. QA-LoRA uses QAT but uses weight-only quantization \cite{xu2023qa}. ParetoQ also applies QAT to weight-only quantization \cite{liu2025paretoq}. BitDistiller is focused on 2 and 3 bit weight models, producing models with unacceptably low accuracy \cite{du2024bitdistiller}.  Finally, BitNet a4.8 uses 4-bit activations and 1-bit weights, but changes activation functions and retrains from scratch, impractical for quantizing existing LLMs \cite{wang2024bitnet}.

\section{Methods}

\subsection{Simple Large Language Model Quantization-Aware Training}\label{sec:simple_recipe}

Our approach is to apply QAT to train LLMs by following three simple practices: 1) add quantization to the model with the straight through estimator, 2) set step sizes through calibration, then refine using LSQ, and 3) train end-to-end using knowledge distillation.  Details for each of these are provided below, including the specific configuration we demonstrate here.

\textbf{Add quantization to the model with the straight through estimator}.  Quantization can be added to any combination of activation, cache, and weight tensors, as dictated by the target deployment platform.  Here, we quantize all three.

For training, we employ scaled quantization, as is common in QAT, to preserve data ranges, thereby making it easier to adapt existing training recipes.  For symmetric quantization, which we employ here, this consists of quantizing (sub)tensor $\bf{x}$ with step size $s$ according to
\begin{equation}
\hat{\mathbf{x}} = \lfloor \min(\max( \mathbf{x} / s, b_l), b_u) \rceil s,
\end{equation}
where $\lfloor . \rceil$ is the round to nearest operator, and $b_l$ and $b_u$ are the lower and upper bound (smallest and largest values) for the corresponding integer at the given precision.  We pass the gradient through the round operation according to the straight through estimator \cite{bengio2013estimating}.  For inference, weights are scaled to integers by dividing by their step size prior to deployment.  Activations are similarly scaled during operation immediately prior to a linear or matrix multiplication layer, with the result then scaled back by multiplying by both associated scales.  For our demonstration, we employ a scale per tensor for activations, and a scale per output channel for weights.

\textbf{Set step sizes through calibration, then refine using LSQ}. We use percentile initialization for activation step sizes.  For our demonstration, we calibrate with 5 batches of 128 samples from the training set, setting the step size to the value at the 99.91, 99.99, and 99.995 percentile for 4-, 8-, and 16-bit activations, respectively.

We introduce a novel approach to set the weight step size to minimize a convex approximation of mean squared error.  For a given set of weights $\bf{w}$, with precision $p$, we approximate the magnitude of the bound of the quantization range as $b=(2^{p-1}-0.5)$.  We assume a weight below this bound, that is $|w| < sb$, is equally likely to fall anywhere in its given quantization bin, treating its quantization error as a uniform random number bounded by $[-\frac{s}{2}, \frac{s}{2}]$, which has an expected squared value of $\frac{s^2}{12}$.  This lets us approximate mean squared error as
\begin{equation}
\hat{\epsilon} = \Sigma_i \max( \tfrac{s^2}{12}, H(|w_i| - sb) (|w_i|-sb)^2)
\end{equation}
where $H$ is the Heaviside step function.  This function is convex with respect to $s$ and so can easily be solved for the step size that approximately minimizes mean squared quantization error.  

Following calibration, we employ the LSQ algorithm to adjust all quantization step sizes during training.  We found that boosting the learning rate for the activation quantization step sizes by a factor of 50 was beneficial.

\textbf{Train end-to-end using knowledge distillation.}  We employ standard end-to-end training using the original model training dataset when it is available, and high quality publicly available data when it is not. Training labels are provided using knowledge distillation, employing the original unquantized model as the teacher.  Mixing a loss derived using teacher provided training labels with a loss using next-token-prediction may be beneficial in some instances, but we, along with others \cite{liu2023llm}, have found this not to be necessary.

We test our method on four 7- or 8-billion parameter models available from Huggingface: meta-llama/Llama-2-7b, meta-llama/Meta-Llama-3-8B, allenai/Llama-3.1-Tulu-3.1-8B, and ibm-granite/granite-3.1-8b-instruct. The Tulu-3 instruction-tuned variant of Llama-3.1-8B was chosen because all instruction tuning datasets are publicly available. We train base models using the publicly available DCLM dataset \cite{li2024datacomp}, and train instruct models using a mixture of the same Supervised Fine-Tuning (SFT) dataset used to finetune the original model (75\%) and the DCLM dataset (25\%). Further training details related to datasets and hyper-parameters are provided in Appendix \ref{sec:app_training}.

\subsection{Precisions tested}

\begin{figure}
\centerline{\includegraphics[width=0.9\columnwidth]{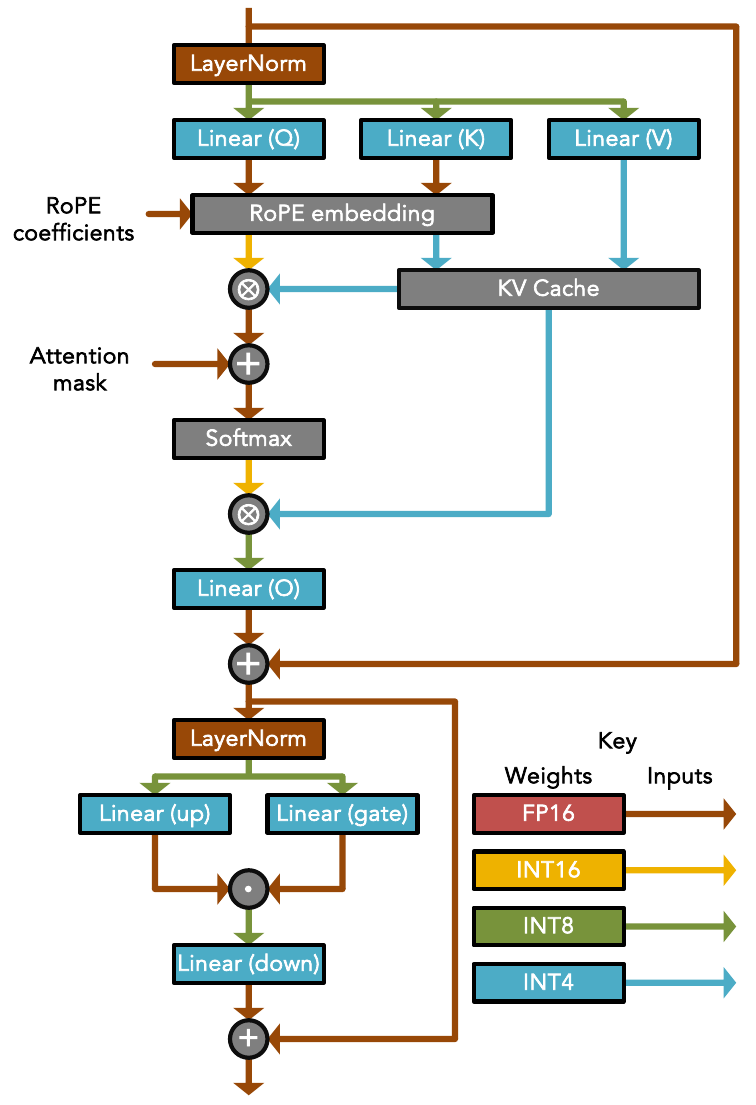}}
 \caption{Transformer block with the A8-C4-W4 precision configuration tested here (A8-C8-W4 also tested) is fully compatible with the NorthPole processor.}
  \label{fig:precision}
\end{figure}

 For our demonstration, we use standard round-to-nearest, uniform, symmetric quantization, with 8-bit activations, 4- or 8-bit KV-Cache, and 4-bit weights, denoted A8-C8-W4 and A8-C4-W4.  We use either token-wise dynamic quantization, or tensor-wise static quantization for all activations, indicated with 'd' and 's', respectively, e.g. A8d-C8-W4.  Figure \ref{fig:precision} shows the precisions used in the attention and multi-layer perceptron blocks used in this study.  The input activations and weights of the final output linear layer are quantized to 8-bits, while the embedding layer remains fp16.  To better match existing LLM quantization papers that do not quantize the matrix multiplication operations, we use INT16 precision for the two non-cache inputs to these layers, the query tensor and the softmax output tensor. Further, as the softmax output tensor is fully encapsulated by the flash attention CUDA kernel, which does not natively support QAT, we do not quantize this tensor during training. This allows us to take advantage of the significant memory and throughput advantages of flash attention \cite{dao2022flashattention}, and empirically has no impact on task accuracy.

\subsection{Model Evaluation}

All evaluations are performed with the latest version of the lm-evaluation-harness \cite{eval-harness}.  For reproducibility, see Appendix \ref{sec:app_eval} for the exact flags used during evaluation.

\subsection{Weight rotation analysis} \label{methods:rotation}

Recent PTQ methods alleviate difficult to quantize outliers by rotating model weight matrices \cite{liu2024spinquant,ashkboos2024quarot}. To understand if our method is learning such rotations, for each model weight matrix, we compare values prior to quantization with those after SpinQuant or QAT (in the case of SpinQuant, using as initial weights the values after folding in LayerNorm scale factors) as follows.  First, we measure the orthogonal Procrustes distance \cite{schonemann1966generalized}, $d_p(A,B)=\min_{R} d_f(RA, B)$ for left-side rotations and $d_p(A,B)=\min_{R} d_f(AR, B)$ for right-side rotations, where $R$ is constrained to the set of rotation matrices and $d_f$ is Frobenius distance, and use whichever is smaller; this metric gives us the \textit{non-rotational distance} between weights pairs, that is, how much of the weight change \textit{can not} be accounted for by matrix rotation.  Next we measure the total distance between weight pairs using Frobenius distance and subtract from this the orthogonal Procrustes distance; this metric provides the \textit{rotational distance}, how much of the weight change \textit{can} be accounted for by matrix rotation.  To facilitate comparison across layers, we normalize all distances by the Frobenius norm of the corresponding original, unquantized weight. Two layers, v\_proj and o\_proj, are rotated twice by SpinQuant, once on the left side and once on the right side, which our measurement is insensitive too, and so they are omitted from our analysis. 

\section{Results}

\subsection{Comparison with PTQ approaches}
We demonstrate that SiLQ leads to significantly higher accuracy than leading PTQ techniques from the literature on the CSR, OLLMv1, and OLLMv2 benchmarks, examining averages across each set (Table \ref{tab:ptq_all_evals}), as well as individual benchmark values (Tables \ref{tab:ptq_csr}, \ref{tab:ptq_ollmv1}, and \ref{tab:ptq_ollmv2} in the appendix).  We run our method for 128,000 steps for these comparisons.

Notably, our approach produces a quantized model that, on some metrics, matches the accuracy of the original full precision model, and across each benchmark average leads to an accuracy gap of no more than 2 absolute percentage points.  Across the board, our method consistently outperforms leading PTQ methods\footnote{Sections \ref{sec:app_spinquant} and \ref{sec:app_smoothquant} in the appendix provide implementation details of SpinQuant and SmoothQuant, respectively.}, often by large margins. This performance advantage is maintained even when comparing PTQ methods employing dynamic quantization to our method employing static quantization, which is simpler to implement and computationally less expensive for both training and inference.

Looking at the impact of training duration, Figure \ref{fig:acc_vs_steps_rel} shows the accuracy of our technique begins to exceed that of SpinQuant on instruction tuned models after only 500 training steps, and continues to improve with longer training.  On the more modern and challenging benchmarks, OLLMv1 and OLLMv2, accuracy continues to improve even at the longest training duration tested, suggesting further training may still improve accuracy.  This curve nicely highlights the ability of QAT to trade-off time-to-solution for accuracy.

Across 128,000 steps of QAT, the model sees 0.1\%, 0.07\%, and 0.08\% of the tokens needed to train the full precision base model for Llama-3-8B, Llama-3.1-Tulu-3.1-8B, Granite-3.1-8B-Instruct, requiring less than 2 weeks on a single, 8-H100 node for the 8B parameter models tested.  Crucially, this demonstrates that our QAT method can quantize a model with near lossless accuracy at the cost of less than a tenth of a percent increase in total model training resources.

\subsection{Comparison with alternative QAT method}

LLM-QAT is the only published QAT method that meets our comparison criteria of quantizing activations, cache, and weights at precisions sufficient to approach full precision accuracy. While our main focus is the more useful instruction-tuned models, as LLM-QAT applies QAT to base models, we apply our method to the Llama-2-7B base model for a direct comparison\footnote{LLM-QAT source code does not support Llama 3 models \cite{llmqatrepo}. Appendix \ref{sec:app_llmqat} provides details about LLM-QAT implementation.}. Table \ref{tab:llm_qat} shows that our approach using the open-source DCLM dataset, on the same model with the same number of training samples as in the LLM-QAT paper, leads to higher accuracy than LLM-QAT in considerably less time.  If we use the time spent by LLM-QAT for generating samples from the model to instead perform further QAT, the accuracy comes close to that of the original model (within 1\% on all banchmarks, and within 0.2\% on OLLMv2, Table \ref{tab:llm_qat}, last row). Using an open-source dataset for training is both simpler and more scalable, and saves considerable effort and computational resources compared to using synthetic data generated from the model as in \cite{liu2023llm}, and can lead to better results.

\begin{table}[h]
\begin{scriptsize}
\begin{center}
\caption{Using the same number of training samples, SiLQ achieves significantly better accuracy in considerably less time than LLM-QAT on Llama-2-7B with A8d-C8-W4 precision. Time is wall-clock time in hours for training plus data generation (only needed for LLM-QAT) on 8 H100 GPUs. With similar total time, our technique improves further, approaching the full precision baseline accuracy. LLM-QAT does not quantize the head.}
\label{tab:llm_qat}
\begin{tabular}{c|cc|ccc}
\hline
\textbf{Method} &\textbf{Hours} &\textbf{Samples}  & \textbf{CSR} & \textbf{OLLM} & \textbf{OLLM} \\ 
&  & \textbf{(thousands)} &  & \textbf{v1} & \textbf{v2}\\
\hline
Baseline & - & - & 64.09 & 50.64 & 8.39  \\
\cdashline{1-5}
LLM-QAT & 104 & 96 & 61.42 & 47.33 & 7.56     \\
\cellcolor{gray!30}SiLQ & \cellcolor{gray!30}1.6  & \cellcolor{gray!30}96 & \cellcolor{gray!30}63.13 & \cellcolor{gray!30}49.08 & \cellcolor{gray!30}7.64  \\       
\cellcolor{gray!30}SiLQ & \cellcolor{gray!30}100  & \cellcolor{gray!30}6,080 & \cellcolor{gray!30}63.37 & \cellcolor{gray!30}50.38 & \cellcolor{gray!30}8.19\\ \hline
\end{tabular}
\end{center}
\end{scriptsize}
\end{table}

\subsection{Generality of our approach}

For simplicity, in the above experiments we use the original SFT dataset for QAT. However, if the original dataset is not available, will QAT with a publicly available substitute suffice? To test this, we repeat the 8,000 step run for the Granite-3.1-8B-Instruct model, replacing the original closed source SFT dataset with the open source Tulu3 SFT data.  QAT with the Tulu3 dataset leads to better scores across all 3 benchmarks (Table \ref{tab:tulu_sft}).  This is not surprising given the quality of the dataset, however, it shows the generality of our QAT approach, since the Tulu3 dataset was constructed to optimize the scores of the Llama models, and suggests that QAT can take advantage of potential accuracy gains afforded by newer datasets.

As a further test, we use the Tulu3 SFT dataset for QAT of the Meta-Llama-3-8B-Instruct model for 8,000 steps. Although we do not have access to the SFT dataset used to train the original model for comparison, our QAT technique with the open source Tulu3 SFT/DCLM dataset mixture performs remarkably well, exceeding the original model's accuracy on two of three benchmarks (Table \ref{tab:tulu_sft}). Based on these results, QAT remains a viable option even in the absence of access to the original closed source SFT datasets, given the availability of good open source alternatives.

\begin{table}
\centering
\caption{High quality open source datasets provide excellent results when used in place of proprietary datasets for QAT.  The open source Tulu3 SFT dataset led to better accuracy than using the model's original training data for Granite-3.1-8B-Instruct.  For Llama-3-8b-Instruct, the original training data is not available for our purposes, but QAT using the Tulu3 SFT dataset is sufficient to approach or exceed the accuracy of the original fp16 model.  For both comparisons, the quantized model was at A8d-C8-W4 precision and underwent 8,000 steps of QAT. Green or red indicate increase or decrease relative to the described baseline.}
\scriptsize
\begin{adjustbox}{max width=\textwidth}
\begin{tabular}{cc|ccc}
\hline
\textbf{Model} &\textbf{SFT Dataset}  &\textbf{CSR} & \textbf{OLLMv1} & \textbf{OLLMv2} \\ 
\hline
\multirow{3}{*}{\shortstack{Granite-3.1-8B\\-Instruct}}  & Original     & 67.45    & 69.45   & 26.54   \\ 

                         & \multirow{2}{*}{Tulu3 SFT}    & 
                         \cellcolor{green!30}67.78    & \cellcolor{green!30}70.30   & \cellcolor{green!30}28.52   \\
& & \cellcolor{green!30}(+0.33) & \cellcolor{green!30}(+0.85) & \cellcolor{green!30}(+1.98) \\
\hline
\multirow{2}{*}{{\shortstack{Llama-3-8B\\-Instruct fp16}}} & \multirow{2}{*}{Original}     & \multirow{2}{*}{63.60}  & \multirow{2}{*}{65.95} & \multirow{2}{*}{24.01}     \\  & & & & \\
\cdashline{1-5}
\multirow{2}{*}{{\shortstack{Llama-3-8B\\-Instruct}}} & \multirow{2}{*}{Tulu3 SFT} & \cellcolor{green!30}63.80 & \cellcolor{green!30}66.13 & \cellcolor{red!30}23.26 \\
& & \cellcolor{green!30}(+0.20)  & \cellcolor{green!30}(+0.18) & \cellcolor{red!30}(-0.75) \\
\hline
\end{tabular}
\label{tab:tulu_sft}
\end{adjustbox}
\end{table}

\subsection{Ablation studies}

To better understand the space around our chosen training and quantization configuration, we perform several ablation studies (Table \ref{tab:ablations}).  For simplicity, we report results for each ablation using a single 8,000 step training run on Granite-3.1-8B-Instruct at A8d-C8-W4 precision. Two factors of those we looked at stood out as having a major influence on accuracy: the use of knowledge distillation, and proper activation calibration.  Specifically, we found that constructing our loss function using knowledge distillation alone lead to much better results than using a mix of knowledge distillation and conventional next token prediction, or next token prediction alone.  This is slightly different from prior work \cite{liu2023llm} showing that a mix of teacher labels and true target labels gave similar results to just using the teacher labels, though the difference may be attributed to their use of a base model vs our use of an instruct model.  Calibrating the step size for activation quantization using the quantile approach led to much better accuracy than using the maximum value from the calibration set.

\begin{table*}
\centering
\caption{Knowledge distillation and quantile activation calibration are critical components of SiLQ at A8d-C8-W4 precision. Results of ablation studies using Granite-3.1-8B-Instruct are shown. The first row is the baseline configuration. For ablations, \textit{KD Ratio} is the ratio of knowledge distillation loss to next token prediction loss, \textit{KD Temp} is the knowledge distillation temperature, \textit{DCLM Ratio} is the ratio of DCLM data to instruct tuning data, \textit{Act Lrx} is the multiplicative scaling factor on learning rate applied to activation quantizer scales, \textit{Act Calib} is the activation quantizer calibration method, \textit{Wgt Calib} is the weight quantizer calibration method, either the MSE based approach introduced here or the calibration method from the LSQ paper, \textit{Online Rot} is whether online rotations were applied as in \cite{ashkboos2024quarot}.  Green or red indicate increase or decrease relative to baseline.}
\scriptsize
\begin{adjustbox}{max width=\textwidth}
\begin{tabular}{ccccccc|cc}
\hline
\textbf{KD} & \textbf{KD} & \textbf{DCLM} & \textbf{Act} & \textbf{Act} & \textbf{Wgt} & \textbf{Online} & \textbf{OLLM} & \textbf{OLLM} \\ 
\textbf{Ratio} & \textbf{Temp} & \textbf{Ratio} & \textbf{Lrx} & \textbf{Calib} & \textbf{Calib} & \textbf{Rot} & \textbf{v1} & \textbf{v2} \\ 
\hline

1.0 & 1.0 & 0.25 & 50 & Quantile & MSE & No & 68.65 & 27.67 \\        
\hline
\cellcolor{gray!30}0.0 & 1.0 & 0.25 & 50 & Quantile & MSE & No & \cellcolor{red!30} 63.36 (-5.29) & \cellcolor{red!30}23.35 (-4.32) \\

\cellcolor{gray!30}0.5 & 1.0 & 0.25 & 50 & Quantile & MSE & No & \cellcolor{red!30} 67.74 (-0.91) & \cellcolor{red!30}25.60 (-2.07) \\

1.0 & \cellcolor{gray!30} 0.5 & 0.25 & 50 & Quantile & MSE & No & \cellcolor{green!30} 68.94 (\texttt{+}0.29) & \cellcolor{red!30}26.33 (-1.34) \\

1.0 & \cellcolor{gray!30} 2.0 & 0.25 & 50 & Quantile & MSE & No & \cellcolor{green!30} 69.72 (\texttt{+}1.07) & \cellcolor{red!30}26.97 (-0.70) \\


1.0 & 1.0 & \cellcolor{gray!30} 0.0 & 50 & Quantile & MSE & No & \cellcolor{green!30} 69.37 (\texttt{+}0.72) & \cellcolor{red!30}26.23 (-1.44) \\

1.0 & 1.0 & \cellcolor{gray!30} 0.5 & 50 & Quantile & MSE & No & \cellcolor{green!30} 68.99 (\texttt{+}0.34) & \cellcolor{red!30}26.90 (-0.77) \\


1.0 & 1.0 & 0.25 & \cellcolor{gray!30} 1 & Quantile & MSE & No & \cellcolor{red!30} 67.69 (-0.96) & \cellcolor{red!30}27.13 (-0.54) \\


1.0 & 1.0 & 0.25 & 50 & \cellcolor{gray!30} Max & MSE & No & \cellcolor{red!30} 63.98 (-4.67) & \cellcolor{red!30}21.01 (-6.66) \\


1.0 & 1.0 & 0.25 & 50 & Quantile & \cellcolor{gray!30} LSQ & No & \cellcolor{red!30} 68.18 (-0.47) & \cellcolor{red!30}27.15 (-0.52) \\


1.0 & 1.0 & 0.25 & 50 & Quantile & MSE & \cellcolor{gray!30} Yes & \cellcolor{green!30} 70.02 (+1.37) & \cellcolor{red!30}27.27 (-0.4) \\

\hline
\end{tabular}
\label{tab:ablations}
\end{adjustbox}
\end{table*}

\subsection{Weight analysis}

\begin{figure}
\begin{center}
\centerline{\includegraphics[width=0.49\textwidth]{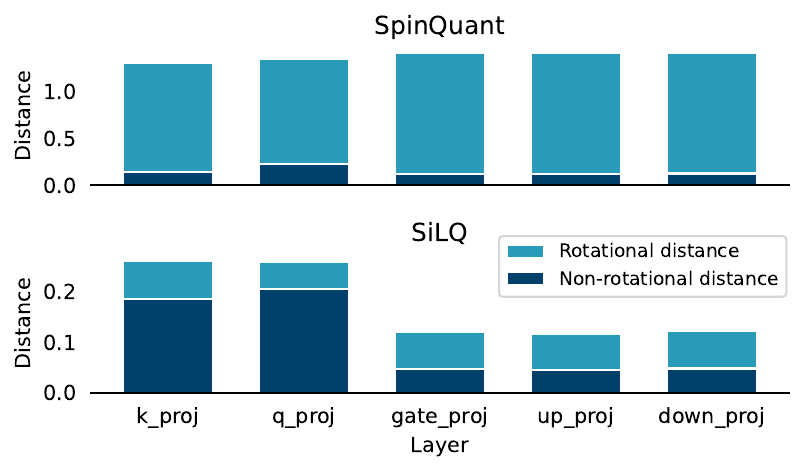}}
 \caption{SiLQ solutions cannot be explained purely by weight matrix rotations; it finds solutions that SpinQuant cannot.  Plotted are weight changes to the LLM linear layers as a result of QAT or SpinQuant factored into the amount explainable by matrix rotation (rotational distance), measured using orthogonal Procrustes distance, and the remainder (non-rotational distance), measured using Frobenius distance minus the orthogonal Procrustes distance.  Changes are measured for each linear layer, normalized by the Frobenius norm of the starting weights, then averaged by layer type.}
  \label{fig:rotation}
\end{center}
\end{figure}

To understand if SiLQ is simply learning weight matrix rotations, and therefore if there is an obvious means to match its performance by improvements to existing rotation based PTQ mehods \cite{liu2024spinquant,ashkboos2024quarot}, we measure how much of the weight change produced by SiLQ can be accounted for by matrix rotation, and perform that same analysis using SpinQuant as a baseline (as described in Section \ref{methods:rotation}).  On the Granite-3.1-8B-instruct A8d-C8-W4 model (Figure \ref{fig:rotation}), we find that rotation accounts for 90\% of all weight changes produced by SpinQuant, with the remainder attributable to the round and clamp operations present in the quantization itself.  In contrast, only 43\% of the weight changes produced by SiLQ can be accounted for by rotation, suggesting that a purely rotational PTQ based approach would be unable to find a similar solution.

\section{Conclusion}
We demonstrate strong performance across three LLM benchmarks on base and instruct models in absolute terms, and relative to leading PTQ and QAT methods, in a commensurate comparison that equalizes quantization constraints.  On Llama3-8B, our approach, quantizing activations, cache, and weights, is able to match the accuracy of the original full precision model on common sense reasoning tasks, while leading PTQ methods that quantize only weights are unable to reach this level of accuracy (for example, Table 3 in \cite{huang2024empirical}) despite the more relaxed quantization requirements. Importantly, we are the first to focus on instruction-tuned models in this space, which are more likely to be deployed than base models.

We found that, even in a case with a relatively small dataset, the longer the QAT duration with knowledge distillation, the higher the accuracy of instruction-tuned models on three extensive benchmarks. This is somewhat surprising, given that further SFT training of the floating point Tulu3 model, for instance, leads to lower accuracy (see Figure 6 in \cite{lambert2024t}), and our best QAT results are obtained with many additional epochs over this dataset with less than one million samples.  Furthermore, contrary to a recent weight-only QAT LLM study which, based on much smaller LLMs (125 million parameters with 100 billion token training budget) found it best to use 10\% of the training budget for QAT \cite{liu2025paretoq}, we show that less than 0.1\% of the budget for QAT can lead to accuracy close to the full-precision counterparts for models with 8-bit activation quantization.  Finally, it is also surprising that, although the instruct models were originally created using a combination of pre-training, SFT, proximal policy optimization, direct preference optimization, and model averaging, QAT with the knowledge distillation loss using only an SFT and pre-training dataset mixture
restores accuracy to nearly the same level as the original model. 

As the field scales up inference-time compute, low latency becomes critical, making these methods increasingly relevant for building performant LLMs capable of reasoning. A simple QAT procedure can, with a relatively short training run on a single 8-H100 node, produce efficient, quantized 8-billion parameter instruction-tuned models that can be deployed with low-power and low-latency on accelerators like NorthPole, enabling new \cite{yao2023tree} and more sustainable AI applications.

\section*{Limitations}

Limitations of our work include training resources, dataset access, specific hardware constraints, specific model sizes, and limited number of trials due to resource constraints.  As with QAT in general, training may take up to 2 weeks for the highest accuracy.  We perform QAT on all model parameters, using 8, 80GB GPUs for the 8B models studied.  
We had access to the IBM internal SFT dataset for our study, however, all other datasets were open source available through HuggingFace, and we conducted experiments to verify that the internal dataset could be replaced by an open-source version.  
All models here were fully quantized to best match the hardware constraints of the NorthPole processor.  Although activations, cache, and weights were fully integer, other operations remained at 16 bits as described in the Methods.
Eight-billion parameter models were the focus of our study; although the method should generalize to larger and smaller models, 8B models are of special interest to us due to their reasonable accuracy and the fact that they can be implemented on the current NorthPole production system.
Finally, due to resource constraints, we chose to use our training budget to study different model variations over different training durations with a single trial per experiment rather than to perform multiple trials on fewer models and durations.  This is common with LLM training, and the clear trends across models justify the decision.


\bibliography{custom}

\appendix
\section{Evaluation Details}\label{sec:app_eval}
We use the following lm-evaluation-harness flags to run each task in the OLLMv1 leader-board and the 8 common-sense reasoning tasks on base models:
\begin{small}
\begin{verbatim}
--model_args pretrained=<model_path>,
      dtype=bfloat16,max_length=4096 \
--tasks <task_name> \
--num_fewshot <num_shots> \
--model hf \
--batch_size auto \
--show_config \
--trust_remote_code
\end{verbatim}
\end{small}
For the OLLMv2 leader-board, the flags are
\begin{small}
\begin{verbatim}
--model_args pretrained=<model_path>,
      dtype=bfloat16 \
--tasks leaderboard \
--model $model_type \ 
--batch_size auto \
--show_config \
--trust_remote_code
\end{verbatim}
\end{small}

For instruction-tuned models, we add the following flags to the above for all tasks:
\begin{small}
\begin{verbatim}
 --apply_chat_template --fewshot_as_multiturn
\end{verbatim}
\end{small}

\begin{table*}
\centering
\caption{Comparison of accuracy on zero-shot CSR tasks with leading PTQ techniques.}
\scriptsize
\begin{adjustbox}{max width=\textwidth}
\begin{tabular}{c|cc|cccccccc}
\hline
\textbf{Model} &\textbf{Bits}  &\textbf{Method} & \textbf{ARC-e} & \textbf{ARC-c} & \textbf{BoolQ} & \textbf{PIQA} & \textbf{SIQA} & \textbf{HellaS.} & \textbf{OBQA} & \textbf{WinoG.} \\ 
\textbf{} & A-C-W  &\textbf{} & \textbf{} & \textbf{} \\ \hline
\begin{tabular}[c]{@{}c@{}}\multirow{4}{*}{Llama-3-8B}\end{tabular}
&\begin{tabular}[c]{@{}c@{}}16-16-16\end{tabular}
         & Baseline  & 77.9 & 53.0  & 81.0 & 81.0 & 46.9 & 79.2 & 45.0 & 72.7 \\ \cdashline{2-11}
& \begin{tabular}[c]{@{}c@{}}\multirow{3}{*}{8d-8-4}\end{tabular}
        & SmoothQuant*  & 63.05 & 38.57 &	72.97 &	74.43 &	43.76 &	70.56 &	39.20 &	67.32  \\
        && SpinQuant  & 74.66 & 47.44 & 72.51 & 78.62 & 45.50 & 76.84 & 43.40 &  72.77 \\        
        && \cellcolor{gray!30}SiLQ &  
        \cellcolor{gray!30}79.88 & 
        \cellcolor{gray!30}52.47 &
        \cellcolor{gray!30}81.44 &
        \cellcolor{gray!30}79.82 &
        \cellcolor{gray!30}46.47 &
        \cellcolor{gray!30}78.26 &
        \cellcolor{gray!30}45.60 &
        \cellcolor{gray!30}73.64  \\ \hline
\begin{tabular}[c]{@{}c@{}}\multirow{4}{*}{\shortstack{Llama-3.1\\-Tulu-3.1-8B}}\end{tabular}
&\begin{tabular}[c]{@{}c@{}}16-16-16\end{tabular} 
         & Baseline  & 81.23 & 55.29 & 85.29 & 80.63 & 55.83 & 79.29 & 49.40 & 72.30 \\ \cdashline{2-11}
&\begin{tabular}[c]{@{}c@{}}\multirow{3}{*}{8d-8-4}\end{tabular} 
        & SmoothQuant*  & 73.06 &	47.35 &	82.39 &	77.15 &	49.95 &	73.47 &	43.60 &	66.61 \\
        && SpinQuant  & 78.24 & 52.56 & 76.88 & 80.90 & 51.54 & 79.31 & 47.20 & 73.16 \\
        && \cellcolor{gray!30}SiLQ &  
        \cellcolor{gray!30}81.23 & 
        \cellcolor{gray!30}54.18 &
        \cellcolor{gray!30}85.38 &
        \cellcolor{gray!30}81.07 &
        \cellcolor{gray!30}55.17 &
        \cellcolor{gray!30}78.74 &
        \cellcolor{gray!30}48.00 &
        \cellcolor{gray!30}72.93 \\ \hline
\begin{tabular}[c]{@{}c@{}}\multirow{9}{*}{\shortstack{Granite-3.1-8B\\-Instruct}}\end{tabular}
&\begin{tabular}[c]{@{}c@{}}16-16-16\end{tabular}
         & Baseline & 75.08 & 53.41  & 88.87 & 78.84& 55.17 & 77.89 & 49.60& 68.82 \\ \cdashline{2-11}
&\begin{tabular}[c]{@{}c@{}}\multirow{3}{*}{8d-8-4}\end{tabular}
        & SmoothQuant*  & 65.82 & 45.90 &	85.78 &	75.08 &	48.36 &	73.17 &	42.00 &	65.35  \\
        && SpinQuant    & 69.49 & 47.53 & 85.99 & 78.67 & 50.97 & 75.78 & 45.80 &  73.48 \\
        && \cellcolor{gray!30}SiLQ &  
        \cellcolor{gray!30}75.04 & 
        \cellcolor{gray!30}53.24 &
        \cellcolor{gray!30}88.41 &
        \cellcolor{gray!30}78.62 &
        \cellcolor{gray!30}54.71 &
        \cellcolor{gray!30}76.87 &
        \cellcolor{gray!30}48.40 &
        \cellcolor{gray!30}68.98 \\ \cdashline{2-11}
&\begin{tabular}[c]{@{}c@{}}\multirow{2}{*}{8s-8-4}\end{tabular} 
        & SmoothQuant*  & 46.89 &	34.47 &	64.71 &	64.58 &	40.23 &	52.60 &	34.80 &	54.22 \\
        && \cellcolor{gray!30}SiLQ &  
        \cellcolor{gray!30}73.48 & 
        \cellcolor{gray!30}52.39 &
        \cellcolor{gray!30}88.17 &
        \cellcolor{gray!30}78.45 &
        \cellcolor{gray!30}54.35 &
        \cellcolor{gray!30}76.86 &
        \cellcolor{gray!30}46.4&
        \cellcolor{gray!30}69.14 \\ \cdashline{2-11}

&\begin{tabular}[c]{@{}c@{}}\multirow{3}{*}{8d-4-4}\end{tabular}
        & SmoothQuant*  & 59.01 &	41.13 &	75.54 &	71.22 &	45.19 &	64.64 &	38.80 &	56.35 \\
        && SpinQuant  & 68.01 & 45.65 & 85.41 & 76.06 & 48.87 & 71.38 & 42.40 &  67.17 \\
        && \cellcolor{gray!30}SiLQ &  
        \cellcolor{gray!30}74.37 & 
        \cellcolor{gray!30}52.90 &
        \cellcolor{gray!30}88.38 &
        \cellcolor{gray!30}78.73 &
        \cellcolor{gray!30}55.12 &
        \cellcolor{gray!30}76.61 &
        \cellcolor{gray!30}47.80 &
        \cellcolor{gray!30}69.46 \\    
        \hline
\multicolumn{3}{l}{*head not quantized} \\

\end{tabular}
\label{tab:ptq_csr}
\end{adjustbox}
\end{table*}

\begin{table*}
\centering
\caption{Comparison of accuracy on OLLMv1 tasks with leading PTQ techniques.}
\scriptsize
\begin{adjustbox}{max width=\textwidth}
\begin{tabular}{c|cc|cccccc}
\hline
\textbf{Model} &\textbf{Bits}  &\textbf{Method} & \textbf{ARC-c} & \textbf{HellaS.} & \textbf{MMLU} & \textbf{TruthfulQA} & \textbf{WinoG.} & \textbf{GSM8K}\\ 
\textbf{} & A-C-W  &\textbf{} & \textbf{} & \textbf{} \\ \hline
\begin{tabular}[c]{@{}c@{}}\multirow{4}{*}{Llama-3-8B}\end{tabular}
&\begin{tabular}[c]{@{}c@{}}16-16-16\end{tabular}
         & Baseline  & 59.04 & 81.97  & 65.34 & 43.95 & 76.80 & 48.82  \\ \cdashline{2-9}
& \begin{tabular}[c]{@{}c@{}}\multirow{3}{*}{8d-8-4}\end{tabular}
        & SmoothQuant*  & 46.08 &	67.63 &	46.34 &	35.85 &	70.09 &	4.93 \\
        && SpinQuant  & 56.14 & 79.13 & 60.99 & 40.97 & 75.14 & 35.56  \\
        && \cellcolor{gray!30}SiLQ &  
        \cellcolor{gray!30}56.48 & 
        \cellcolor{gray!30}81.14 &
        \cellcolor{gray!30}63.79 &
        \cellcolor{gray!30}44.20 &
        \cellcolor{gray!30}76.16 &
        \cellcolor{gray!30}45.03 \\ \hline
\begin{tabular}[c]{@{}c@{}}\multirow{4}{*}{\shortstack{Llama-3.1\\-Tulu-3.1-8B}}\end{tabular}
&\begin{tabular}[c]{@{}c@{}}16-16-16\end{tabular} 
         & Baseline  & 57.85 & 81.84 & 61.87 & 59.86 & 74.66 & 81.27  \\ \cdashline{2-9}
&\begin{tabular}[c]{@{}c@{}}\multirow{3}{*}{8d-8-4}\end{tabular} 
        & SmoothQuant*  & 49.49 &	74.51 &	47.50 &	54.69 &	71.19 &	51.33 \\
        && SpinQuant  & 55.55 & 81.15 & 58.39 & 52.81 & 74.35 & 79.23  \\
        && \cellcolor{gray!30}SiLQ &  
        \cellcolor{gray!30}59.64 & 
        \cellcolor{gray!30}81.36 &
        \cellcolor{gray!30}62.36 &
        \cellcolor{gray!30}58.20 &
        \cellcolor{gray!30}74.27 &
        \cellcolor{gray!30}82.94  \\ \hline
\begin{tabular}[c]{@{}c@{}}\multirow{9}{*}{\shortstack{Granite-3.1-8B\\-Instruct}}\end{tabular}
&\begin{tabular}[c]{@{}c@{}}16-16-16\end{tabular}
         & Baseline & 62.62 & 84.48  & 65.34 & 66.23 & 75.37 & 73.84  \\ \cdashline{2-9}
&\begin{tabular}[c]{@{}c@{}}\multirow{3}{*}{8d-8-4}\end{tabular}
        & SmoothQuant*  & 57.59 &	78.96 &	55.96 &	63.25 &	73.24 &	40.94 \\
        && SpinQuant    & 58.70 & 79.79 & 58.35 & 59.74 & 78.77 & 57.77  \\
        && \cellcolor{gray!30}SiLQ &  
        \cellcolor{gray!30}66.98 & 
        \cellcolor{gray!30}84.15 &
        \cellcolor{gray!30}65.46 &
        \cellcolor{gray!30}64.66 &
        \cellcolor{gray!30}76.16 &
        \cellcolor{gray!30}71.49 \\ \cdashline{2-9}
&\begin{tabular}[c]{@{}c@{}}\multirow{2}{*}{8s-8-4}\end{tabular} 
        & SmoothQuant*  & 27.65 &	40.31 &	27.04 &	55.32 &	50.20 &	10.92 \\
        && \cellcolor{gray!30}SiLQ &  
        \cellcolor{gray!30}65.87 & 
        \cellcolor{gray!30}84.01 &
        \cellcolor{gray!30}64.7 &
        \cellcolor{gray!30}65.61 &
        \cellcolor{gray!30}74.74 &
        \cellcolor{gray!30}70.2 \\ \cdashline{2-9}
&\begin{tabular}[c]{@{}c@{}}\multirow{3}{*}{8d-4-4}\end{tabular}
        & SmoothQuant*  & 38.99 &	59.98 &	28.96 &	53.10 &	54.93 &	2.73 \\
         && SpinQuant  & 54.69 & 74.80 & 44.26 & 57.55 & 72.69 & 36.01 \\
         && \cellcolor{gray!30}SiLQ &  
        \cellcolor{gray!30}65.44 & 
        \cellcolor{gray!30}84.00 &
        \cellcolor{gray!30}65.04 &
        \cellcolor{gray!30}65.81 &
        \cellcolor{gray!30}74.90 &
        \cellcolor{gray!30}70.36 \\ \hline
\multicolumn{3}{l}{*head not quantized} \\ 

\end{tabular}
\label{tab:ptq_ollmv1}
\end{adjustbox}
\end{table*}

\begin{table*}
\centering
\caption{Comparison of accuracy on OLLMv2 tasks with leading PTQ techniques.}
\scriptsize
\begin{adjustbox}{max width=\textwidth}
\begin{tabular}{c|cc|cccccc}
\hline
\textbf{Model} &\textbf{Bits}  &\textbf{Method} & \textbf{BBH} & \textbf{GPQA} & \textbf{IFEval} & \textbf{MATH} & \textbf{MMLU-Pro} & \textbf{MUSR}\\ 
\textbf{} & A-C-W  &\textbf{} & \textbf{} & \textbf{} \\ \hline
\begin{tabular}[c]{@{}c@{}}\multirow{4}{*}{Llama-3-8B}\end{tabular}
&\begin{tabular}[c]{@{}c@{}}16-16-16\end{tabular}
         & Baseline  & 23.92 & 7.61  & 15.64 & 4.46 & 25.01 & 5.55  \\ \cdashline{2-9}
& \begin{tabular}[c]{@{}c@{}}\multirow{3}{*}{8d-8-4}\end{tabular}
        & SmoothQuant*  & 6.10 &	1.68 &	16.95 &	3.20 &	6.53 &	5.46 \\
        && SpinQuant  & 21.12 & 2.85 & 15.31 & 3.78 & 21.31 & 9.28  \\
        && \cellcolor{gray!30}SiLQ &  
        \cellcolor{gray!30}22.67 & 
        \cellcolor{gray!30}6.15 &
        \cellcolor{gray!30}14.33 &
        \cellcolor{gray!30}4.31 &
        \cellcolor{gray!30}21.90 &
        \cellcolor{gray!30}6.61 \\ \hline
\begin{tabular}[c]{@{}c@{}}\multirow{4}{*}{\shortstack{Llama-3.1\\-Tulu-3.1-8B}}\end{tabular}
&\begin{tabular}[c]{@{}c@{}}16-16-16\end{tabular} 
         & Baseline  & 18.25 & 2.74 & 82.40 & 22.36 & 21.11 & 11.82  \\ \cdashline{2-9}
&\begin{tabular}[c]{@{}c@{}}\multirow{3}{*}{8d-8-4}\end{tabular} 
        & SmoothQuant*  & 11.70 &	1.34 &	76.22 &	19.26 &	9.82 &	4.60 \\
        && SpinQuant  & 17.53 & 3.02 & 78.44 & 16.99 & 16.84 & 11.37  \\
        && \cellcolor{gray!30}SiLQ &  
        \cellcolor{gray!30}20.60 & 
        \cellcolor{gray!30}1.73 &
        \cellcolor{gray!30}82.80 &
        \cellcolor{gray!30}23.64 &
        \cellcolor{gray!30}21.11 &
        \cellcolor{gray!30}12.71  \\ \hline
\begin{tabular}[c]{@{}c@{}}\multirow{9}{*}{\shortstack{Granite-3.1-8B\\-Instruct}}\end{tabular}
&\begin{tabular}[c]{@{}c@{}}16-16-16\end{tabular}
         & Baseline & 33.84 & 8.72  & 69.80 & 21.22 & 28.14 & 17.75  \\ \cdashline{2-9}
&\begin{tabular}[c]{@{}c@{}}\multirow{3}{*}{8d-8-4}\end{tabular}
        & SmoothQuant*  & 20.55 &	5.70 &	49.17 &	17.90 &	17.33 &	9.79 \\
        && SpinQuant  & 25.37 & 7.94 & 54.95 & 6.57 & 18.90 & 14.38  \\
        && \cellcolor{gray!30}SiLQ &  
        \cellcolor{gray!30}33.12 & 
        \cellcolor{gray!30}8.05 &
        \cellcolor{gray!30}70.19 &
        \cellcolor{gray!30}18.81 &
        \cellcolor{gray!30}28.40 &
        \cellcolor{gray!30}16.28 \\ \cdashline{2-9}
&\begin{tabular}[c]{@{}c@{}}\multirow{2}{*}{8s-8-4}\end{tabular} 
        & SmoothQuant*  & 4.20 &	2.96 &	39.51 &	3.71 &	1.62 &	1.60 \\
        && \cellcolor{gray!30}SiLQ &  
        \cellcolor{gray!30}33.18 & 
        \cellcolor{gray!30}9.17 &
        \cellcolor{gray!30}68.7 &
        \cellcolor{gray!30}18.2 &
        \cellcolor{gray!30}27.66 &
        \cellcolor{gray!30}17.24 \\ \cdashline{2-9}
&\begin{tabular}[c]{@{}c@{}}\multirow{3}{*}{8d-4-4}\end{tabular}
        & SmoothQuant*  & 5.13 &	0.06 &	29.45 &	0.64 &	1.50 &	6.10 \\
         && SpinQuant  & 13.43 & 3.80 & 46.41 & 2.64 & 9.25 & 11.27 \\
         && \cellcolor{gray!30}SiLQ &  
        \cellcolor{gray!30}33.15 & 
        \cellcolor{gray!30}9.40 &
        \cellcolor{gray!30}70.29 &
        \cellcolor{gray!30}18.05 &
        \cellcolor{gray!30}27.43 &
        \cellcolor{gray!30}16.43 \\ \hline
\multicolumn{3}{l}{*head not quantized} \\          

\end{tabular}
\label{tab:ptq_ollmv2}
\end{adjustbox}
\end{table*}

\section{Training Details}\label{sec:app_training}
As described in section \ref{sec:simple_recipe}, during QAT, base models are trained using the publicly available DCLM dataset and instruct models are trained using a mixture of SFT data and DCLM. For Llama-3.1-Tulu-3.1-8B we use the tulu-3-sft-mixture \cite{lambert2024t}, and for Granite-3.1-8b-instruct\footnote{https://huggingface.co/ibm-granite/granite-3.1-8b-instruct}, we use the same SFT dataset used to train the original model. For Llama models, we found it necessary during training to use the LlamaTokenizerFast tokenizer class  which removes the bos\_token following the SpinQuant source code.

To train, we use SFTTrainer \cite{sfttrainer} with the AdamW optimizer (beta1=0.9, beta2=0.95, epsilon=1e-10, bfloat16) \cite{loshchilov2017decoupled}, employ cross entropy loss with labels derived from the knowledge distillation teacher at a temperature of 1, disable dropout to avoid adverse affects on knowledge distillation, use a base learning rate of 5e-6 for 8,000 steps with a cosine learning rate schedule and a minimum learning rate set to 10\% of the initial value, no warm-up period, and use a weight decay of 0.1. The learning rate was the best from \{2e-6, 5e-6, 1e-5\} at 8,000 training steps for both Llama-3-8B and granite-3.1-8b-instruct. The learning rate is reduced for longer training runs by the inverse square-root of the increase in training steps, e.g. when increasing training steps by a factor of 4, the learning rate is reduced to half \cite{shen2024power}. Similarly, for shorter runs, the learning rate is increased by the square-root of the decrease in steps. We train with a batch size of 128 with sequence length 1024, without packing on a single 8-H100 GPU node. 

\section{SpinQuant Comparison}\label{sec:app_spinquant}
We compare our technique on all models with SpinQuant, the leading state-of-the-art PTQ method for 4-bit weights and 8-bit activations at the time of this publication.  For a fair comparison, we run the SpinQuant author's code\footnote{https://github.com/facebookresearch/SpinQuant} on the original bfloat16 publicly available models.  To avoid the additional inference overhead due to online rotations, we modify the code to support the no online Hadamard rotations option described in the paper.  We use the default parameters from the paper with the following changes to match the hardware constraints of our target deployment platform.  We use symmetric quantization for the activations with no groups for most activations in the network. We use group quantization for only the key cache and not the value cache, with group size 128, as the extra key cache group-wise scales can be integrated more readily into the matrix multiply operation on chip. The final head linear layer inputs and weights are also quantized to 8-bits.  

\section{SmoothQuant Comparison}\label{sec:app_smoothquant}
For an additional comparison, we also compare against SmoothQuant for all models.  We use the authors' original code modified to use 4-bit weights instead of 8, and to control the precision of the KV-cache at either 4- or 8-bits while leaving the output of the query linear layer unconstrained to more closely match SpinQuant and our QAT precision settings.  We chose alpha=0.4 as the best among {0.2, 0.3, ... 0.7} for the Granite-3.1-8b-instruct A8d-C8-W4 model, and used the same value for the remaining models. Note that the final head linear layer remains unconstrained in their code.

\section{LLM-QAT Comparison}\label{sec:app_llmqat}
We use the original code from the authors for comparison using default parameter settings for both data generation and training \cite{llmqatrepo}. Note that the LLM head is not quantized in their code. We train for 750 steps for our method in order to compare with the same number of samples, 96,000 as in \cite{liu2023llm}, and adjust the learning rate to 1.63e-5 accordingly.  We perform a second run training for 47,500 steps at learning rate 2e-6 to approximately match the total wall clock time used by LLM-QAT for data generation and training.  We set the max sequence length to 2048 to match LLM-QAT. We compare our method with LLM-QAT only on llama-2-7b, because the code provided does not support Llama-3 models.
\end{document}